\documentclass[11pt]{article}

\usepackage[preprint]{acl}

\usepackage{times}
\usepackage{latexsym}

\usepackage[T1]{fontenc}

\usepackage[utf8]{inputenc}

\usepackage{microtype}

\usepackage{inconsolata}

\usepackage{booktabs}
\usepackage{graphicx}
\usepackage[most]{tcolorbox}

\usepackage{bbm}
\usepackage{multirow}
\usepackage{array}
\usepackage{xspace}
\usepackage{bbding}
\usepackage{makecell}
\usepackage{enumitem}
\usepackage{pifont}

\title{DMC-CF: Dynamic Multimodal CounterFactual QA benchmark for Causal Reasoning}

\author{Junzhe Zhang\textsuperscript{1}, 
Huixuan Zhang\textsuperscript{1}, 
Guirong Wang\textsuperscript{2},
Xingyao Zhang\textsuperscript{2},\\
\textbf{Pei Liu\textsuperscript{2},
Lin Qu\textsuperscript{2},
HU WEI\textsuperscript{2},
Xiaojun Wan\textsuperscript{1}}\\ 
\textsuperscript{1}Wangxuan Institute of Computer Technology, Peking University \\
\textsuperscript{2}Alibaba Group\\
\{junzhezhang, zhanghuixuan\}@stu.pku.edu.cn  \\ wanxiaojun@pku.edu.cn \\
\{wangguirong.wgr, kongwang\}@alibaba-inc.com \\
\{senwen.zxy, xumu.xu, xide.ql\}@taobao.com }

\begin{document}
\maketitle
\begin{abstract}
With the rapid advancement of multimodal large language models (MLLMs), models have demonstrated increasingly powerful multimodal capabilities. However, whether MLLMs trained through statistical learning can truly understand the causal relationships underlying the real world remains a key research question. In recent years, numerous multimodal causal reasoning datasets have been proposed. Nevertheless, these datasets are either limited in scale or constructed from synthetic images and videos, cartoon-based content, or other non-realistic multimodal sources.
To address these limitations, we collect real-world videos and construct DMC-CF-Static, a large-scale benchmark for multimodal causal counterfactual reasoning. Furthermore, to mitigate issues such as data contamination in traditional static evaluation, we represent causal events using causal graphs and propose the Dynamic Graph Intervention (DGI) framework to build the dynamic evaluation benchmark DMC-CF-Dynamic from DMC-CF-Static. Experimental results on the overall DMC-CF, which includes both static and dynamic evaluation benchmarks, demonstrate that the multimodal causal reasoning capabilities of current multimodal large language models in real-world scenarios still require substantial improvement.



\end{abstract}

\section{Introduction}
Recent advances\cite{gemmateam2025gemma3technicalreport, comanici2025gemini25pushingfrontier, singh2026openaigpt5card} in multimodal large models have significantly improved the ability to process and understand information across multi modalities. Among these multimodal tasks, multimodal causal reasoning has emerged as an important problem, requiring models to understand causal relationships across multimodal events. In real world, events are often determined by multiple causal factors, and the ability to identify the underlying causal relationships of real-world events has become an important criterion for evaluating whether models trained through statistical learning can truly understand the world.


\begin{figure}[t]
    \centering
    \includegraphics[width=\columnwidth]{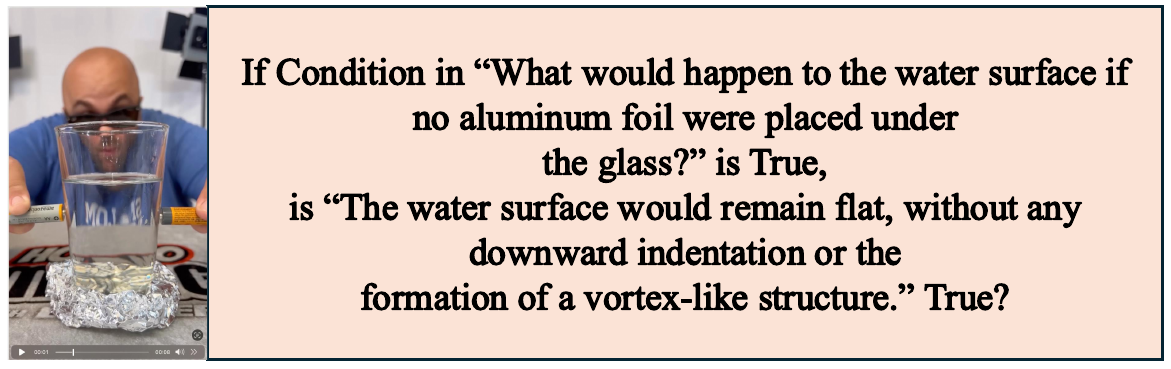}
    \caption{An example of Causal CounterFactual Video QA.}
    \label{fig:intro_case}
\end{figure}

In recent years, a growing number of studies on multimodal causal reasoning have been proposed\cite{li-etal-2025-multimodal-causal, foss2025causalvqaphysicallygroundedcausal, NEURIPS2024_a8955806, NEURIPS2024_5f280960}.
These datasets can be categorized into two types based on their sources. One type is constructed from non-real multimodal data, such as artificially generated scenes or cartoon-style environments, while the other is built by sampling from real-world data.
However, datasets constructed from non-realistic data often exhibit a certain degree of distributional discrepancy from the real world, while existing datasets built from real-world data are generally limited in scale. For example, CausalVQA\cite{foss2025causalvqaphysicallygroundedcausal} contains only 1,786 samples. 

Therefore, in order to better evaluate causal reasoning abilities of models in real-world scenarios, we collected suitable real-world video and manually annotated a total of approximately 5k multimodal causal reasoning samples to construct our static multimodal causal reasoning dataset, DMC-CF-static, encompassing 6 different categories of video. Besides ensuring the authenticity of video sources and high annotation quality, it also provides fine-grained annotation metadata.



Further, static benchmarks often suffer from the risk of explicit or implicit data contamination. The evaluation results on traditional static datasets may not comprehensively reflect models’ causal reasoning capabilities in real-world scenarios.
To address this issue, dynamic evaluation\cite{yang2025dynamicmultimodalevaluationflexible,liu2025reasoningmultimodallargelanguage,zhang2026kbedmedynamicmultimodalevaluation} has recently been proposed as a new evaluation paradigm to mitigate the limitations of traditional static benchmarks.
However, existing multimodal dynamic evaluation methods either support only limited ranges of dynamic variation or rely on knowledge bases and other external knowledge sources to obtain the updated gold answers. In contrast, causal reasoning in real-world scenarios is often highly complex, making it difficult to develop a system capable of automatically generating correct gold answers for dynamically modified questions.

When revisiting traditional studies on causal reasoning, we observe that the concept of intervention experiments inherently embodies a dynamic perspective. Intervention experiments analyze whether a factor has a causal relationship with an observed outcome by dynamically intervening on that factor.

\begin{table*}[htp]
    \centering
    \setlength{\tabcolsep}{1.2mm}{
    \renewcommand\arraystretch{1.1}
    \begin{tabular}{ccccccc}
    \toprule 
    \multirow{2}*{Benchmark} & \multirow{2}*{V/I} & Video/Image              & Video/Image          & Video/Image    & QA-Pair     & Dynamic    \\
                             &             & Source                   & Type                 & Number         & Number      & Property     \\\hline 
    MuCR                     & I       & Synthetic                & Multiple             & 12k            & 12k         & Static     \\ 
    CausalVQA                & V       & Real                     & Physical             & 779            & 0.8k        & Static     \\
    Causalchaos!             & V       & Cartoon                  & Multiple             & 161 episodes   & 5k            & Static   \\
    DMC-CF                   & V       & Real                     & Multiple             & 1614           & 5k(S)+10k(D)      & Static and Dynamic     \\
    \bottomrule
    \end{tabular}}
    \caption{Comparison of DMC-CF and previous researches of mainstream Multimodal Causal Reasoning Benchmark. V/I indicates whether the multimodal information modality of the dataset is video or image. S is for Static, D is for Dynamic.}
\label{table:comparison}
\end{table*}

Motivated by this insight, we first introduce the definition of causal graphs and leverage high-quality manually annotated metadata to represent the causal relationships in each sample using a corresponding static causal graph. Subsequently, inspired by the dynamic nature of intervention experiments, we propose the Dynamic Graph Intervention (DGI), which captures dynamical causal intervention, and further construct the dynamic evaluation benchmark DMC-CF-dynamic based on DMC-CF-static.

For each static counterfactual QA in DMC-CF-static, we construct a corresponding intervention causal graph based on the manually annotated counterfactual questions. By identifying intervention entry points that satisfy predefined requirements within the intervention causal graph, we select the intervention entry point to conduct new intervention to construct dynamic QAs. Meanwhile, the gold answer is inferred through reasoning over the causal graph structure, thereby ensuring the quality of the dynamically generated Video Causal Reasoning QA pairs. We also generate different versions of dynamic samples under multiple settings for test.


Experimental results on the overall DMC-CF, which includes both static and dynamic evaluation benchmarks, demonstrate that the multimodal causal reasoning capabilities of current multimodal large language models in real-world scenarios still require substantial improvement. Our contributions are\footnote{The code and dataset will be released upon publication.}:
\begin{itemize}

    \item We propose DMC-CF-static, a multimodal causal reasoning dataset constructed through high-quality manual annotations on real-world videos. The dataset contains more than statistic 5k multimodal causal counterfactual QA pairs spanning 6 categories, serving as a benchmark for evaluating models’ causal reasoning capabilities in real-world scenarios.

    \item We introduce causal graphs to represent the causal relationships among events and further employ Dynamic Graph Intervention(DGI) to dynamically construct 10k causal counterfactual QA pairs, thereby establishing the dynamic evaluation benchmark DMC-CF-Dynamic.

    \item Extensive experiments demonstrate that current mainstream multimodal large language models are capable of successfully answering relatively simple multimodal causal counterfactual QA questions. However, their accuracy remains generally low when dealing with complex dynamically generated counterfactual QA.


\end{itemize}













\section{Related Works}

\subsection{Multimodal Causality Reasoning}
Recently, a growing number of studies\cite{li2023intentqa, wu-etal-2023-acquired, Li_2022_CVPR} on causal reasoning have emerged in the multimodal domain.
MuCR \cite{li-etal-2025-multimodal-causal} uses synthetic images and text pairs to evaluate MLLMs. It further provides a comprehensive assessment of MLLM performance at the image, phrase, and sentence levels.
CausalVQA \cite{foss2025causalvqaphysicallygroundedcausal}, based on real-world videos, constructs a benchmark containing five types of causal data to evaluate models’ causal reasoning capabilities in physical-world scenarios.
Causalchaos! \cite{NEURIPS2024_a8955806} constructs Why-QA questions using videos derived from cartoon-based sources to evaluate models’ video causal QA capabilities. The work further employs causal chains to build more complex causal questions, thereby increasing the difficulty of the benchmark. 
HourVideo \cite{NEURIPS2024_5f280960} is a benchmark designed to evaluate visual-language understanding over hour-long videos. It contains 500 multimodal causal reasoning samples constructed from real-world videos.


However, the aforementioned works either do not rely on real-world videos or are limited in scale. Therefore, we propose DMC-CF counterfactual QA Benchmark.
The main differences between DMC-CF and previous works are summarized in Table \ref{table:comparison}.















\subsection{Dynamic Evaluation}
To address data contamination and saturation issues, recent studies have investigated dynamic evaluation methods \citep{jiang2025raising, yang2025dynamicmultimodalevaluationflexible}, where test samples are either perturbed \citep{yang2025dynamicmultimodalevaluationflexible} or regenerated \citep{jiang2025raising} to adjust task difficulty and minimize contamination effects. In the domain of text-only dynamic evaluation, DyVal \citep{zhu2024dyvaldynamicevaluationlarge} dynamically generates test instances to alleviate data contamination.



However, dynamic evaluation in the multimodal domain is still relatively underexplored. VLB \citep{yang2025dynamicmultimodalevaluationflexible} is among the earliest works to jointly bootstrap images and text by modifying image objects or backgrounds, rephrasing or replacing words in questions, and inserting related or unrelated textual information to perturb original VQA pairs. 
Although perturbation-based approaches can diversify test inputs, their effects are generally more limited than methods that regenerate entirely new evaluation data.
\cite{zhang2026kbedmedynamicmultimodalevaluation} Leveraging multimodal knowledge to dynamically extend existing static multimodal benchmarks and control their difficulty levels.
However, existing multimodal dynamic evaluation methods either support only limited ranges of dynamic variation in multimodal video causal reasoning tasks or are entirely incompatible with such settings. Considering that the concept of intervention experiments in causal reasoning is inherently dynamic, we are inspired to construct a multimodal causal counterfactual benchmark capable of supporting dynamic evaluation.



\begin{figure}[t]
    \centering
    \includegraphics[width=0.9\columnwidth]{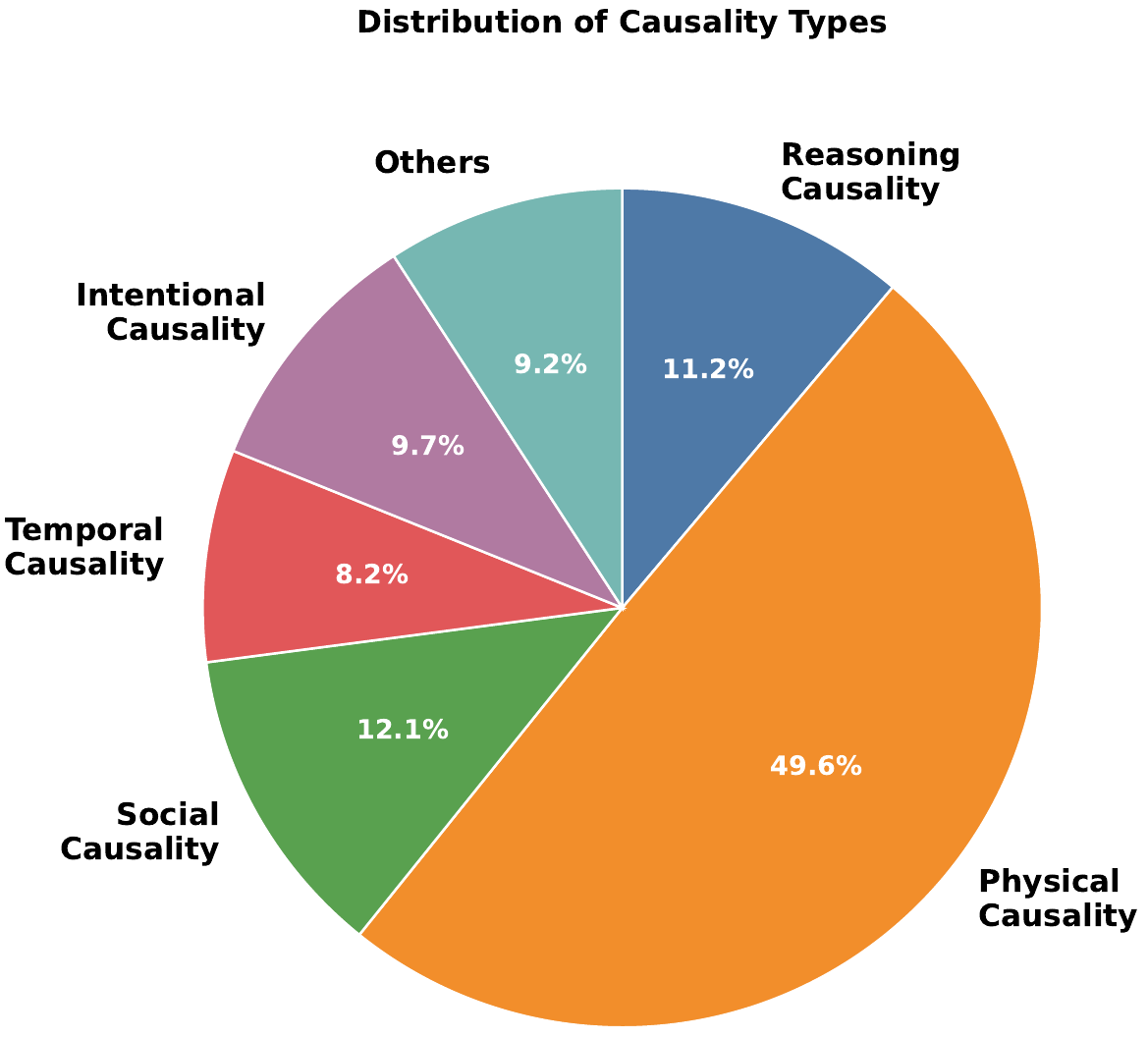}
    \caption{Distribution of Causality Types.}
    \label{fig:causality_distribution}
\end{figure}

\begin{figure*}[t]
    \centering
    \includegraphics[width=0.88\textwidth]{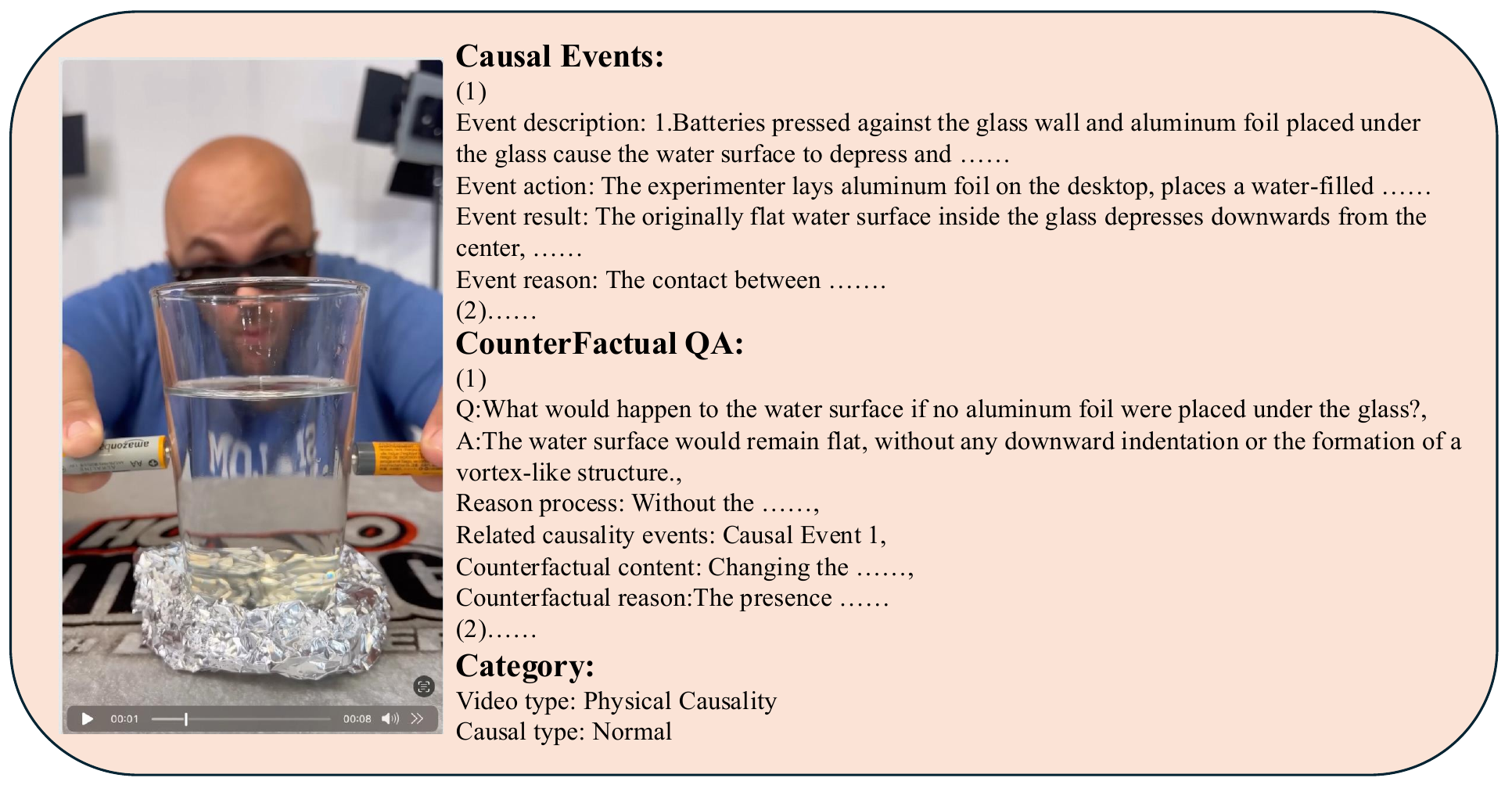}
    \caption{An example of our annotated sample.}
    \label{fig:case_study}
\end{figure*}

\section{DMC-CF-Static}
\subsection{Video Collection}
We first manually select real-world videos from public accessible video streaming platform such as YouTube that satisfy our following proposed requirements. We require the videos to originate from real-world sources rather than AI-generated or cartoon-based content. In addition, the videos must depict continuous events without excessive editing or multiple cuts, and they must contain clear and concrete causal events, with every object involved in the events being explicitly identifiable. The causal events contained in the collected video clips ultimately span 6 major categories of causal types, including Reasoning Causality, Physical Causality, Social Causality, Temporal Causality, Intentional Causality and Others. The details of required criteria can be seen in Appendix \ref{sec:appendix:human annotation}. After extensive collection and rigorous filtering, we ultimately gathered 1614 distinct videos that satisfy our requirements. 



\subsection{Data Annotation}
Annotators are first asked to identify the video segments corresponding to causal events and annotate the associated metadata, such as video descriptions and lists of causal events. They are then required to label the start and end timestamps corresponding to each causal event in the video.

We further instruct annotators to annotate multiple intervention hypotheses and their corresponding causal counterfactual questions for each video. In total, we annotate 5317 counterfactual questions, with an average of 3.3 counterfactual questions per video. 
We specifically identify two relatively challenging categories of data in the dataset according to predefined criteria: multi-object causal reasoning samples and long reasoning chain samples. 
We categorize samples involving more than four primary objects and less than four causal reaction chains as multi-object causal data, resulting in 202 videos that satisfy this criterion. Samples involving less than five primary objects but causal reaction chains longer than three are categorized as long-chain causal data, with 487 videos meeting this requirement. An annotation example can be seen in Figure \ref{fig:case_study}. For the convenience of unified testing and evaluation, we convert the annotated QA pairs into a Yes/No QA (YNQA) format. We named these static counterfactual video YNQAs as \textbf{L1\_Y} subset. An example is shown in Figure \ref{fig:intro_case}. 







\subsection{Data Statistics}
The distribution across different categories is illustrated in the Figure \ref{fig:causality_distribution}. The details of statistics are presented in Table \ref{tab:data_statistics}.
The average duration of the video clips is 8.87 seconds, while the manually annotated counterfactual questions contain an average of 53.18 words.

\begin{table}[t]
\centering
\begin{tabular}{lc}
\toprule
\textbf{Statistic} & \textbf{Value} \\
\midrule
Number of videos & 1614 \\
Number of QA pairs & 5317 \\
Average video length(s) & 8.87 \\
Average question words &  53.18 \\
Average QA pairs per video & 3.3 \\
\bottomrule
\end{tabular}
\caption{Statistics of the DMC-CF-Static dataset.}
\label{tab:data_statistics}
\end{table}



\begin{figure*}[t]
    \centering
    \includegraphics[width=0.85\textwidth]{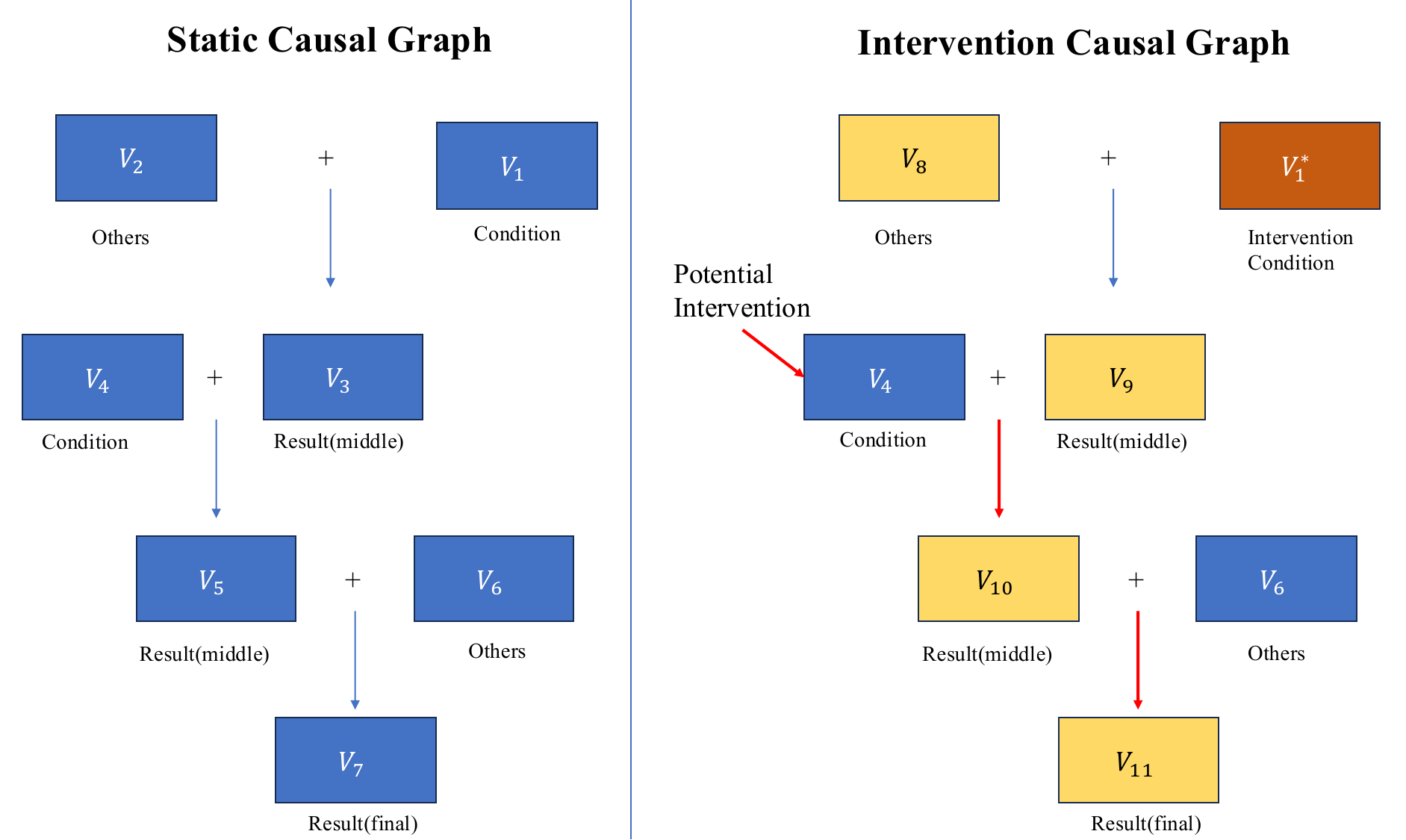}
    \caption{An example of Causal Graph. The left side of the figure illustrates the static causal graph corresponding to the static data, while the right side shows the intervention causal graph corresponding to the manually annotated counterfactual question.}
    \label{fig:Dynamic Causal Graph}
\end{figure*}


\section{DMC-CF-Dynamic}
\subsection{Graph Definition}
We introduce causal graphs composed of different types of nodes to represent the causal events within videos, with an overview illustrated in the Figure \ref{fig:Dynamic Causal Graph}. Specifically, we categorize the nodes into Condition nodes, Result(middle) nodes, Result(final) nodes, and Others nodes.
Condition nodes represent the input condition, which can be manipulated to perform intervention experiments.
Result nodes represent the outcome inferred from the corresponding input conditions. If a result node is not the final outcome of the causal event, it is regarded as an intermediate inferred result and is therefore categorized as a Result(middle) node. In contrast, if a result node corresponds to the final outcome of the causal event depicted in the video clip, it is defined as a Result(final) node.
Other nodes such as reasoning rules or background knowledge are categorized as Others type. The primary motivation for distinguishing Condition nodes from Others nodes is that we aim to conduct intervention experiments by modifying input conditions rather than altering world knowledge or physical laws.



\subsection{Dynamic Graph Intervention}
For each counterfactual QA sample in DMC-CF-Static, we first apply a MLLM(referred as DGM) to extract Causal Graph from the video, namely Static Causal Graph.
We then apply DGM to construct a corresponding causal graph based on the associated video content, annotated metadata, and counterfactual QA pair in DMC-CF-Static, namely Intervention Causal Graph. A counterfactual question typically arises from intervening on the specific node, thereby forming the counterfactual assumption.
Using the manually annotated metadata and counterfactual assumptions, we 
construct the intervention causal graph corresponding to each counterfactual QA. We preserve the node in the original static causal graph if a node remains valid in the intervention causal graph.
The overview of intervention causal graph constructed from the manually annotated counterfactual question is illustrated on the right side of the Figure \ref{fig:Dynamic Causal Graph}, ${{V_{1}}}^*$ is the corresponding intervention node. We refer to such intervention node as ${{V_{I_1}}}^*$.

After constructing the intervention causal graph, we further identify potential suitable intervention entry points for secondary dynamic interventions. Specifically, we define each intervention entry point as a triplet $((V_{I_2}, V_{RM}, V_{RF}))$. Here, $V_{I_2}$(e.g., $V_4$) denotes a Condition-type node potentailly applicable for secondary intervention, $V_{RM}$(e.g., $V_{10}$) represents an Result(middle) node, and $V_{RF}$(e.g., $V_{11}$) corresponds to the Result(final) node associated with the corresponding manually annotated counterfactual QA in DMC-CF-Static for the current video clip.

In the intervention causal graph, for each potential intervention entry point $(V_{I_2}, V_{RM}, V_{RF})$, we require $V_{I_2}$ to be one of the predecessor nodes of $V_{RM}$, and $V_{RM}$ to be one of the predecessor nodes of $V_{RF}$. In other words, our assumption is that intervening on the suitable Condition node $V_{I_2}$ invalidates the intermediate result $V_{RM}$, which further causes the final result $V_{RF}$ to no longer hold. Here, $V_{I_2}$ cannot be the first intervention node ${V_{I_1}}^*$(e.g., ${V_{1}}^*$) corresponding to the original counterfactual assumption in the manually annotated static counterfactual question. Meanwhile, $V_{RF}$ must correspond to the result node associated with the original manually annotated counterfactual question.



\begin{table*}[htp]
    \centering
    \setlength{\tabcolsep}{1.5mm}{
    \renewcommand\arraystretch{1.1}
    \begin{tabular}{ccccccccccc}
    \toprule 
    \multirow{2}*{Method} & \multicolumn{7}{c}{ACC}                  & \multicolumn{3}{c}{F1}   \\ \cmidrule(r){2-8} \cmidrule(l){9-11}
                          & L1\_N & L1\_Y & L2\_N & L2\_Y & L1      & L2         & All              & L1    & L2    & All    \\\hline 
    GPT-5.2                 & 90.52 & 59.92 &	49.31 & 61.16 &	75.19   & 55.20      & 68.90            & 74.85 & 54.88 & 68.78  \\ 
    Claude-4              & 85.02 & 72.61 &	46.43 &	69.94 & 78.82   & 58.14      & 72.34            & 78.54 & 57.14 & 72.05  \\
    Gemini-3.1-pro        & 88.18 & 86.31 & 53.10 & 73.68 & 87.25   & 63.33      & \textbf{79.72}   & 87.28 & 62.56 & \textbf{79.57}  \\\hline  
    Qwen3VL-Instruct        & 92.35 & 42.75 & 68.12 & 28.52 & 67.54   & 48.44      & 61.53            & 65.42 & 46.27 & 59.32  \\ 
    Qwen3VL-Thinking        & 80.18 &	84.44 &	39.03 &	71.02 & 82.31   & 54.93      & \textbf{73.69}   & 82.30 & 53.80 & \textbf{73.58}  \\
    \bottomrule
    \end{tabular}}
    \caption{Main Results of tested models on L1 and L2 Causal CounterFactual samples. 
    }
\label{table:main_result}
\end{table*}

\subsection{Graph-based QA Construction}
Once the intervention causal graph and the intervention entry point $(V_{I_2}, V_{RM}, V_{RF})$ are obtained, we only need to generate an intervention condition on $V_{I_2}$ that causes $V_{RM}$ to no longer hold. We employ the DGM to generate a new intervention condition that does not conflict with the original counterfactual assumption, thereby obtaining the second intervened Condition node ${V_{I_2}}^*$(e.g., ${V_4}^*$).
Based on this, we can construct causal counterfactual QA with deterministic gold answers, specifically, ask whether $V_{RF}$ holds when both the original intervention assumption${V_{I_1}}^*$(e.g., ${V_1}^*$) and the second intervention assumption corresponding to ${V_{I_2}}^*$ (e.g., ${V_4}^*$) are simultaneously satisfied. Furthermore, we use DGM to integrate the original intervention assumption with the intervention assumption of (${V_{I_2}}^*$), making the generated question more natural and coherent.
According to our design, new intervention ${V_{I_2}}^*$ will cause conclusion $V_{RF}$ to no longer hold, so the gold label is N.
We refer to these data samples as \textbf{L2\_N} subset.

Meanwhile, we also construct the (\textbf{L2\_Y}) subset based on the intervention causal graph. Specifically, we replace the second intervention assumption with the content of another randomly selected Condition node $V_{rdC}$ in the original intervention causal graph, excluding the first intervention node itself. Under this setting, the counterfactual question becomes whether $V_{RF}$ still holds when both the first counterfactual assumption ${V_{I_1}}^*$ and the assumption corresponding to $V_{rdC}$ are satisfied. In this case, the gold label is Y.

The original manually annotated counterfactual questions ask whether the corresponding counterfactual outcome would still hold under a given counterfactual assumption. We refer to this type of counterfactual question as \textbf{L1\_Y}.
Following a similar idea, we replace the original counterfactual assumption with the content of a randomly selected original Condition node from the static causal graph, while still asking whether the counterfactual outcome would hold under this modified condition. In this setting, the gold answer becomes N, thereby enabling the construction of the \textbf{L1\_N} subset.
In summary, by jointly leveraging static causal graphs and intervention causal graphs, we perform different intervention experiments through modifying the input counterfactual assumptions, thereby generating questions that ask whether the same counterfactual outcomes in the static dataset still hold under different intervention conditions. To answer these questions correctly, models are required to possess strong multimodal causal reasoning capabilities, enabling them to infer whether the same conclusion is True under different counterfactual assumptions introduced by different intervention experiments.






\section{Experiments}
\subsection{Experimental Setup}
We primarily evaluate several closed-source models, including OpenAI’s GPT-5\cite{singh2026openaigpt5card}
, Gemini-3.1-pro\cite{gemini3pro_modelcard_2025}, and Claude-4\cite{anthropic_claude4_systemcard_2025}, as well as two open-source models, Qwen3VL-235B-A22B-Thinking(referred as Qwen3VL-Thinking) and Qwen3VL-235B-A22B-Instruct(referred as Qwen3VL-Instruct)\cite{bai2025qwen3vltechnicalreport}.
Among these models, Gemini and the Qwen series support native video input. Therefore, during evaluation, we directly provide the counterfactual QA pairs together with the corresponding video files. In contrast, GPT and Claude do not support native video processing. For these models, following default settings, we perform frame extraction on the video files by default at intervals of every 2 seconds, with a maximum of 16 frames sampled. For longer video clips, frames are uniformly sampled across the entire clip. The QA pairs together with the sampled frames are then provided as input to GPT and Claude for evaluation.

We apply Accuracy and Macro-F1 as the evaluation metrics. We calculate macro-F1 with sklearn toolkit.
Specifically, we report the answer accuracy of the models on four different subsets, the L1 and L2 counterfactual QA sets, as well as the entire dataset. In addition, we compute the Macro-F1 scores on the L1 and L2 counterfactual QA sets and the overall dataset.


\subsection{Main Results}
The results presented in the Table \ref{table:main_result} indicate that there is still substantial room for improvement for current mainstream open-source and closed-source multimodal large language models on the DMC-CF dataset. For MLLMs equipped with explicit thinking or reasoning capabilities, both the answer accuracy and F1 scores on L1 questions are relatively strong. However, their performance drops notably on the more challenging L2 questions, where both accuracy and F1 scores remain relatively low.

In particular, these models perform worst on the L2\_N subset. Even the strongest closed-source multimodal models achieve only around 50\% accuracy, while the performance of open-source models is even lower. These findings suggest that current MLLMs can handle relatively shallow reasoning required for L1 questions, but struggle substantially when faced with counterfactual questions involving multiple counterfactual assumptions. In such cases, the models are required to internally simulate the corresponding intervention causal graph and further reason about the outcomes of additional interventions on the modified causal graph. Under these more complex settings, the causal reasoning capabilities of current models deteriorate considerably. It can be observed that among the closed-source multimodal large language models, Gemini, which natively supports video input, demonstrates the strongest multimodal causal reasoning capability.



Meanwhile, we also evaluate both the Thinking and Instruct variants of the open-source Qwen3VL models. As expected, the Thinking variant significantly outperforms the Instruct variant overall. However, the results also reveal an interesting observation. The Instruct version of Qwen achieves the highest accuracy among all evaluated models on the L1\_N and L2\_N subsets, which is counterintuitive. Intuitively, one would expect models with deeper reasoning capabilities to perform better, yet the model without explicit deep reasoning attains higher accuracy on these subsets.

Based on our analysis of the dataset construction process, we hypothesize that this phenomenon may arise from the fact that the reasoning process of the Instruct model is relatively shallow, causing it to rely on implicit decision heuristics when answering questions. For example, once the input hypothesis conflicts with the video content, the model may directly output “N” for the queried result without performing deeper causal reasoning. This observation suggests that evaluating multimodal causal reasoning capabilities on DMC-CF requires jointly considering the results across all subsets, rather than relying on performance on individual subsets alone.


\subsection{Ablation of DGM}
\begin{table*}[t]
\centering
\setlength{\tabcolsep}{1.8mm}{
\renewcommand\arraystretch{1.1}
\begin{tabular}{lcccccc}
\hline
Subset & DGM                      & GPT-5.2 & Claude-4 & Gemini-3.1-pro & Qwen3VL-Ins & Qwen3VL-Thi \\
\hline
L2           & Gemini-3.1-pro     & 55.22	& 58.99    & 62.07          & 48.03       & 53.63\\
L2           & Qwen3VL-Thi        & 50.45   & 55.08    & 56.14	        &48.26        &51.83\\
L1\_Y        & /                  & 58.45   & 69.40    & 85.69          & 40.76       & 86.28 \\
\hline
\end{tabular}}
\caption{Model performance(ACC) of applying Qwen3VL-Thinking as DGM in subsets of L2. We also report performance on corresponding static L1\_Y samples. Qwen3VL-Ins denotes Qwen3VL-Instruct, Qwen3VL-Thi denotes Qwen3VL-Thinking.}
\label{tab:ablation of dgm}
\end{table*}


We use the Thinking variant of Qwen3VL to generate L2 questions in order to evaluate the performance of different multimodal large language models when used as the DGM. Results can be seen in Table \ref{tab:ablation of dgm}, and we also show the performance on the L1\_Y(which is static) samples. It can be observed that, for tested models, the performance differences across counterfactual QA generated using different models as the DGM are relatively minor. 
Overall, we recommend evaluating models using multiple dynamically generated data constructed with different DGM, while also jointly considering their performance on the static benchmark.



\subsection{Performance on different Categories}

\begin{table}[t]
\centering
\begin{tabular}{lccc}
\toprule
\textbf{Model} & \textbf{Others} & \textbf{MO} & \textbf{LC}\\
\midrule
GPT-5.2          & 70.52 & 70.19 & 64.72 \\
Claude-4         & 74.12 & 71.80 & 68.46 \\
Gemini-3.1-pro         & \textbf{81.03} & \textbf{79.25} & \textbf{76.87} \\\hline 
Qwen3VL-Instruct & 62.51 & 62.99 & 58.74 \\
Qwen3VL-Thinking & \textbf{75.88} & \textbf{72.70} & \textbf{69.02} \\
\bottomrule
\end{tabular}
\caption{Accuracy comparison on different categories, MO means Multi-Object, LC means Long-Chain.}
\label{tab:easy_hard_acc}
\end{table}

We analyze the model performance on different category of data. The results in the table \ref{tab:easy_hard_acc} show that the multi-object causal subset is slightly more challenging than the standard causal reasoning data, while the long-chain causal subset is notably more difficult than other causal reasoning samples. Moreover, Gemini and Qwen3-Thinking continue to achieve the best performance. Based on the current experimental results, increasing the reasoning chain length of causal reasoning data poses a greater challenge to models than increasing the number of objects involved in the causal reasoning process.

\begin{figure}[t]
    \centering
    \includegraphics[width=0.85\columnwidth]{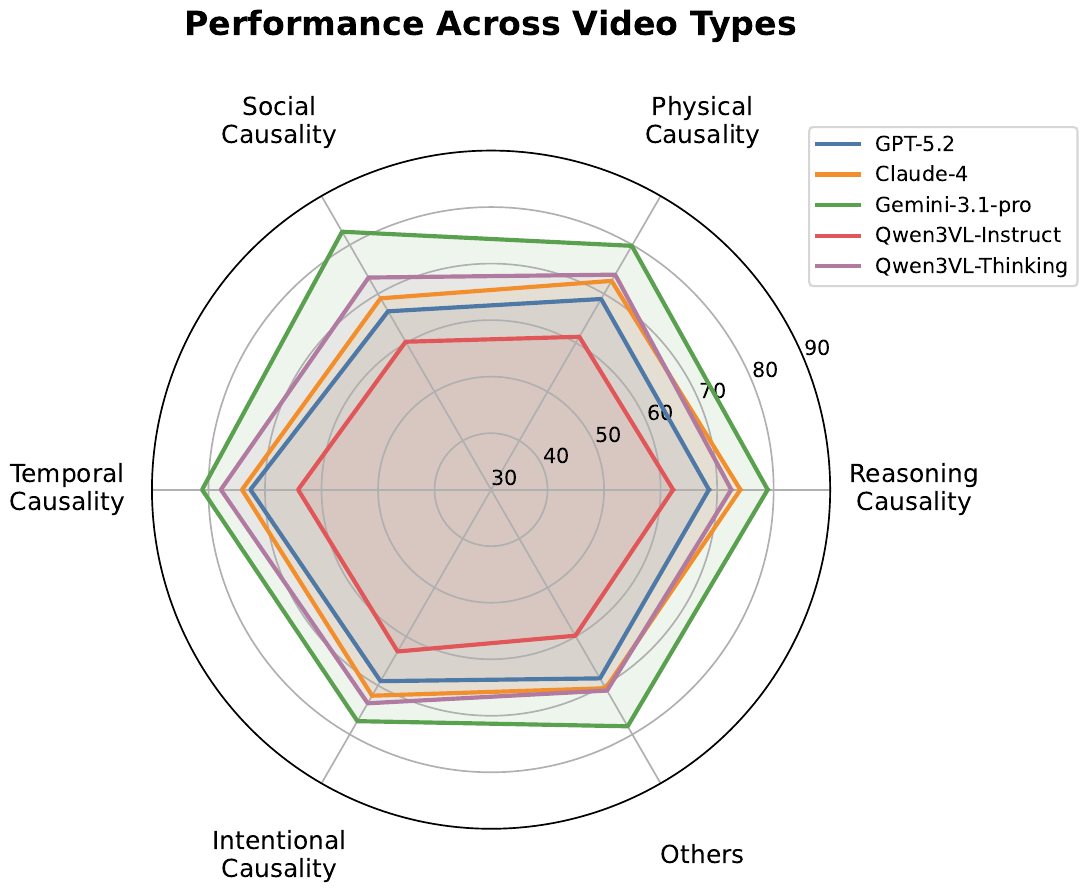}
    \caption{Model performance across video types.}
    \label{fig:category_acc}
\end{figure}

We further analyze the accuracy of five different models across six video causality categories in DMC-CF. As shown in Figure \ref{fig:category_acc}, Gemini-3.1-pro achieves the highest accuracy in all categories, demonstrating the strongest overall multimodal causal reasoning capability. The second-best overall performance is achieved by Qwen3VL-Thinking, which is consistent with the results reported in Table \ref{table:main_result}. It can also be observed that, except for Gemini-3.1-pro, all other models achieve relatively lower accuracy on Social Causality compared with their performance on other causality categories. In addition, closed-source multimodal models with native video understanding capabilities generally demonstrate stronger performance. Models equipped with explicit thinking or reasoning mechanisms also consistently outperform their instruction-only counterparts on counterfactual QA tasks across all video categories.






\subsection{Data Quality Human Verification}
We randomly select 50 samples from the L2\_N subset to conduct quality evaluation along two dimensions: Degree of Dependence on Video and Irrelevant Hallucinations. Degree of Dependence on Video measures the extent to which answering a QA pair requires information from the video, while Irrelevant Hallucinations refer to hallucinated content unrelated to the video.
Notably, due to the presence of counterfactual assumptions, the QA pairs may contain content that is not directly shown in the video. However, if the QA includes information completely unrelated to the video content, we still categorize such content as hallucinations. The detailed human annotation guidelines are provided in the Appendix \ref{sec:appendix:human annotation}. 
The average score of the Degree of Dependence on Video is 3.2, while the average score of Irrelevant Hallucinations is 4.1. This indicates that the dynamically generated test samples contain fewer irrelevant hallucinations and require models to jointly leverage both video information and background knowledge in order to produce accurate answers.



%





\section{Conclusion}
We propose DMC-CF, a Multimodal CounterFactual QA benchmark for Causal Reasoning. Its static component, DMC-CF-Static, consists of video counterfactual QA data obtained through manual collection of real-world videos and high-quality human annotation. We further introduce causal graphs to represent causal events in videos and propose the Dynamic Graph Intervention framework to dynamically construct the dynamic component DMC-CF-Dynamic.
The overall dataset contains 1.6k real-world videos and 15k multimodal causal counterfactual QA samples. We evaluate both open-source and closed-source MLLMs on DMC-CF, and the experimental results demonstrate that the multimodal causal reasoning capabilities of current multimodal large language models in real-world scenarios still require substantial improvement.



\section*{Limitations}
Current work still has some limitations. First, although we attempted to cover diverse causality categories and scenarios, the overall dataset scale and diversity remain limited. Some complex or long-tail causal phenomena are insufficiently represented, which may restrict the evaluation of model generalization in open-domain, cross-domain, or more sophisticated causal reasoning tasks. In addition, the data sources and distributions may still contain certain biases. Future work could further expand the dataset with larger-scale, more diverse, multilingual, and more realistic causal scenarios to improve its coverage and representativeness.


\section*{Ethical Considerations}
Our data is collected from publicly available web sources, and we carefully review each video clip to exclude potentially private, sensitive, or harmful content. 
All annotators are qualified and fairly compensated. 
The annotated counterfactual questions are carefully designed and do not contain potentially harmful content. We use existing models in accordance with their intended use and licensing terms. LLMs are used to assist with writing. We comply with all required licenses during both data collection and model evaluation. Annotators were recruited through a commercial outsourcing workflow and compensated according to standard market practices and local norms.




\bibliography{custom}

\appendix

\section{Human Annotation Details}

\label{sec:appendix:human annotation}

\subsection{Human Annotation Guidelines}
\begin{figure*}[htbp]
    \centering
    \begin{tcolorbox}[title=Video Collection Guideline]
        I. \textbf{Collection Objectives}:
        
            Collect real-world filmed videos on YouTube that meet the following criteria.
        
        II. \textbf{Video Content Requirements}
        
        \begin{itemize}
            \item Videos must feature real people or real-world scenes; AI-generated footage is not allowed.
            \item  The main subject must exhibit motion or action changes and must not remain static for extended periods.
            \item The visual composition should be simple, with clean backgrounds whenever possible (e.g., solid-color walls or minimal scenes), and clutter should be avoided.
            \item Video clips involving causal events should contain minimal editing and maintain continuity.
            \item The video must demonstrate a clear causal relationship.
            \item Resolution: $\geq$720P (1080P recommended)
            \item Frame rate: 30 fps recommended
        \end{itemize}

        \textbf{Multi-clip Instructions}:
        
        A single video may contain multiple clips marked as source data, provided that each corresponding clip contains clear and intuitive causal events.
        
        \textbf{Screening Rules}:
        \begin{enumerate}
            \item  Objects involved in the causal event must be clearly identifiable in the video and referable through language (e.g., wooden blocks, wooden boards, etc., should correspond accurately to visible objects in the video). If multiple identical objects exist and cannot be unambiguously referred to or distinguished linguistically, the video should not pass screening.
            \item The causal relationship should be intuitive, and the causal event should be easy to describe. Click to view counterexamples (e.g., in the referenced video, the coin collision can only be explained through force analysis, and the angles are difficult to describe clearly).
            \item  Content involving gore, violence, soldiers, or other sensitive material must not appear.
            \item Prefer videos that depict the cause and result of a specific event/action, rather than simple production or workflow processes.
            \item Videos must be understandable without relying on subtitles. Do not select videos that require subtitles for comprehension.
        \end{enumerate}

        III. \textbf{Scene Categories}
        
        3.1 Physical Causality
        Collisions and Reactions / Gravity and Motion / Fluid Dynamics / Friction and Resistance / Mechanical Devices
        
        3.2 Social Causality
        Emotional Reactions / Group Behavior / Conflict and Reconciliation / Social Norms
        
        3.3 Intentional Causality
        Goal Setting and Execution / Planning and Implementation / Decisions and Consequences / Obstacles and Overcoming / Multi-step Planning
        
        3.4 Temporal Causality
        Sequential Dependency / Time Constraints / Accumulation Effects / Delayed Consequences / Temporal Reversal
        
        3.5 Reasoning Causality
        Complex Reasoning / Multi-step Causality / Implicit Relationships / Counterintuitive Causality / Hypothesis and Verification
    \end{tcolorbox}
    \caption{Video collection guideline.}
\end{figure*}

\begin{figure*}[htbp]
    \centering
    \begin{tcolorbox}[title=Video Annotation Guideline]
        \textbf{Video Record Documentation}
        
        Each video entry must include the following fields:
        
        a. Video Type:
        Specify the scene category to which the video belongs.
        
        b. URL:
        Provide the complete video link.
        
        c. Account:
        Copy the account name displayed in the lower-left corner of the video exactly as shown.
        
        d. Hashtag:
        Click the three dots in the upper-right corner of the video to ``Description,'' then copy all blue-text hashtags beginning with \#. If none are present, write “None.”
        
        e. Description:
        Copy the original English text from the description page.
        
        f. Introduction:
        Copy the original English text from the introduction page.
        
        g. Start\_time / End\_time:
        Provide the start and end timestamps of the causal event clip.
        
        Format example: 0:15 / 2:30
        
        h. Video Description:
        Use text to describe the causal event process presented in the video.
        
        i. List of Causal Events:
        For videos with a single causal relation, provide 1 causal event; for videos with multiple causal relations, provide 2–3 causal events.
        
        If a video contains multiple causal-event clips, they may be annotated separately, with corresponding timestamps provided.
        
        Causal events should be described as thoroughly as possible, including the action, the result, and the implied causal relationship.
        
        Example: A man pushes a wooden block to the right, causing the block to begin moving to the right due to the applied force.
        
        Action: The man pushes the wooden block to the right.
        
        Result: The wooden block moves to the right.
        
        Causal Relationship: The wooden block begins moving to the right because a rightward force is applied to it.
        
        List the specific contents included in each causal event using the following format:
        
        Causal Event:         Action:        Result:        Causal Relationship:
        
        j. Counterfactual Questions:
        Construct counterfactual QA pairs based on the list of causal events, following the requirements below:
        
        For each causal event, construct 3 counterfactual questions. Each counterfactual question must involve the “cause” of the original causal event. Include both the counterfactual question (Q) and answer (A). Each QA pair must be numbered. Use the following format:

        Q:     A:        Reasoning Process:
        
        Related Causal Event(s): Causal Event 2
        
        Counterfactual Assumption: The wooden board is perfectly smooth.
        
        Manifestation of the Counterfactual: In Causal Event 2, the wooden board is not perfectly smooth.
        
        Q:        A:
    
        k. Counterfactual Questions (English Version):
        Rewrite all counterfactual QA pairs from section j into English while preserving the same format.
        
        l. Multiple-Choice Format:
        Rewrite each counterfactual question in section j as a four-option multiple-choice question, following the requirements below:

        The numbering must correspond to the numbering of the original QA pairs. Each question must include: Question (Q) + Options A/B/C/D + Correct Answer + Reasoning Process. Separate different multiple-choice questions using ***. Use the following format:

        Q: ……         A. ……        B. ……        C. ……        D. ……        Correct Answer: X        Reasoning Process: ……
        
    \end{tcolorbox}
    \caption{Video annotation guideline.}
\end{figure*}

\subsection{Human Quality Evaluation Guideline}

\begin{figure*}[htbp]
    \centering
    \begin{tcolorbox}[title=Video Counterfactual QA Human Quality Evaluation Guideline]

    \paragraph{Evaluation Objective}
    
    For each Video Counterfactual QA sample, conduct a human quality evaluation across the following dimensions:
    \begin{enumerate}
        \item The degree to which answering the question depends on video information
        \item Whether the counterfactual QA contains irrelevant hallucinations
    \end{enumerate}
    
    \textit{Degree of Dependence on Video Information for Answering the Counterfactual QA (Score: 1–5)}

    \begin{enumerate}
        \item 1: Almost no dependence on video information, The question can be answered entirely based on common sense without watching the video.
        \item 2: Slight dependence on video information, Basic knowledge of the objects or scene in the video is required, but the answer still mainly relies on common sense.
        \item 3: Moderate dependence on video information, Understanding part of the behaviors, events, or relationships in the video is necessary.
        \item 4: High dependence on video information, A relatively deep understanding of event progression, character interactions, or action logic in the video is required.
        \item 5: Complete dependence on video information, The question can only be answered by watching and fully understanding the video.
    \end{enumerate}

    \textit{Whether the Counterfactual QA Contains Irrelevant Hallucinations (Score: 1–5)}

    Evaluate whether the counterfactual question and answer are constructed based on actual content in the video, or whether they introduce objects, actions, events, or relationships that do not exist in the video.

    The counterfactual scenario itself is allowed to differ from the real video. Therefore: ``Did not actually happen'' is not necessarily ``Hallucination'', Reasonable counterfactual modifications should not be considered hallucinations. The primary focus of this dimension is whether the QA is grounded in the video content rather than whether the event actually occurred.

    \begin{enumerate}
        \item 1: Severe Hallucination: The question or answer is largely disconnected from the video content, introducing objects, actions, or events that are completely absent from the video. There is little to no meaningful connection to the video.
        \item 2: Major Hallucination: Some content is related to the video, but the QA contains clearly fabricated or incorrect information. There are substantial unsupported inferences.
        \item 3: Minor Hallucination / Partially Ungrounded: The QA is generally related to the video, but includes some unsupported inferences or extrapolations. There is slight over-interpretation or speculation.
        \item 4: Mostly Grounded: The question and answer are largely based on the video content, with only very minor imprecision or ambiguous inference.
        \item 5: Fully Grounded: The counterfactual QA is constructed strictly based on actual video content. It only introduces reasonable modifications to existing events in the video and contains no unsupported fabrication.
    \end{enumerate}
    \end{tcolorbox}
    \caption{Video counterfactual QA quality evaluation guideline}
\end{figure*}

\section{Prompts of DGM}
\label{sec:appendix:dgm prompts}

\begin{figure*}[htbp]
    \centering
    \begin{tcolorbox}[title=Static Causality Graph Extraction Prompt]
        Below is a series of generated node information. Please generate the corresponding node information for video causal reasoning according to the given node information structure.Here is an example of the generated node information structure:
        V1: 
        content: The block has an initial velocity to the right. 
        previous\_condition: None. 
        state: Condition.
        V2: 
        content: The rough wooden board exerts a frictional force on the block opposite to its direction of motion. 
        previous\_condition: None. 
        state: Rule.
        V3: 
        content: Under the action of friction, the block slows down and slides to the right on the wooden board. 
        previous\_condition: V1, V2. 
        state: Result(middle).
        V4: 
        content: The wooden board is long enough. 
        previous\_condition: None. 
        state: Condition.
        V5: 
        content: The block slows down, slides to the right on the wooden board, and eventually stops on the board. 
        previous\_condition: V3, V4. 
        state: Result(final).
        The new input information is as follows: 
        The video is the given input video, 
        and the corresponding causal event annotations are as follows:
        \{causal\_event\_description\}
        causal\_event\_action:\{causal\_event\_action\}
        causal\_event\_result:\{causal\_event\_result\}
        causal\_event\_reason:\{causal\_event\_reason\}
        
    \end{tcolorbox}
    \caption{Static causality graph extraction prompt.}
\end{figure*}

\begin{figure*}[htbp]
    \centering
    \begin{tcolorbox}[title=Intervention Causality Graph Extraction Prompt]
        Below is the generated causal reasoning node graph. Please modify the corresponding reasoning graph according to the counterfactual question information. Use Vx* to denote the intervention node applied to node Vx, and use new numbering for any new nodes generated by the intervention. If an original node still remains valid, retain it:

        The corresponding video is the input video,
        The corresponding causal event information and counterfactual question information are as follows:
        Causal events:
        \{causal\_event\_description\}
        causal\_event\_action:\{causal\_event\_action\}
        causal\_event\_result:\{causal\_event\_result\}
        causal\_event\_reason:\{causal\_event\_reason\}
        ......
        
        Counterfactual question information:
        Q: \{CF\_Q\}
        A: \{CF\_A\}
        Reason\_process: \{reason\_process\}
        Related\_causal\_events: \{related\_causal\_event\}
        Counterfactual\_content: \{CF\_content\}
        Counterfactual\_reason: \{counterfactual\_reason\}
        ......
        
        Please output the corresponding modified causal graph according to the required node format above:
                
    \end{tcolorbox}
    \caption{Intervention causality graph extraction prompt.}
\end{figure*}

\begin{figure*}[htbp]
    \centering
    \begin{tcolorbox}[title=Prompt for Generating $L2\_N$]
        We conducted a counterfactual intervention experiment based on the video. 
    Here, we provide the video clip and the corresponding causal graph extracted after the intervention. Please perform the required intervention experiment and generate a counterfactual question according to the instructions.
    Video: the corresponding input video.
    Causal graph: \{node\_str\}
    Original counterfactual question information: \{previous\_CF\_inf\_str\}
    Intervention experiment requirement: intervene on \{condition\_node\} so that the original \{middle\_node\} no longer holds. The new intervention content must not conflict with the previous counterfactual assumptions.
    New Counterfactual question requirement: generate a question asking whether ``\{res\_node\}'' still holds under both the previous intervention assumptions and the new intervention assumptions. The answer should only require Y/N.
    The counterfactual assumptions used in the intervention experiment must not create shortcuts; that is, the model should not be able to determine whether the related event holds solely based on the intervention assumptions, but instead must combine the video information with reasoning to make the judgment.
    Please output only the generated counterfactual question. Do not include node numbering information in the question.
                
    \end{tcolorbox}
    \caption{Prompt for generating $L2\_N$.}
\end{figure*}

\begin{figure*}[htbp]
    \centering
    \begin{tcolorbox}[title=Prompt for Generating $L2\_Y$]
        We conducted a counterfactual intervention experiment based on the video. 
        Here, we provide the video clip and the corresponding causal graph extracted after the intervention. Please perform the required intervention experiment and generate a counterfactual question according to the instructions.
        Video: the corresponding input video.
        Causal graph: \{node\_str\}
        Original counterfactual question information: \{previous\_CF\_inf\_str\}
        Intervention experiment requirement: assuming the original counterfactual assumptions hold, additionally assume that the condition in \{L1\_ori\_condition\} holds. The new intervention content must not conflict with the previous counterfactual assumptions.
        Counterfactual question requirement: generate a question asking whether ``\{res\_node\}'' holds under both the previous intervention assumptions and the new intervention assumptions. The answer should only require Y/N.
        The counterfactual assumptions used in the intervention experiment must not create shortcuts; that is, the model should not be able to determine whether the related event holds solely based on the intervention assumptions, but instead must combine the video information with reasoning to make the judgment.
        Please output only the generated counterfactual question. Do not include node numbering information in the question, and do not use phrases such as ``as in the video''.
                
    \end{tcolorbox}
    \caption{Prompt for generating $L2\_Y$.}
\end{figure*}

\end{document}